\documentclass[conference]{IEEEtran}
\IEEEoverridecommandlockouts
\usepackage{cite}
\usepackage{amsmath,amssymb,amsfonts}
\usepackage{graphicx}
\usepackage{textcomp}
\usepackage{xcolor}
\usepackage{multicol}
\usepackage{enumitem}
\usepackage{algorithm}
\usepackage{algpseudocode}
\usepackage{makecell}
\usepackage{threeparttable}
\usepackage[bottom]{footmisc}
\usepackage{caption}
\usepackage{booktabs}
\usepackage{threeparttable}
\usepackage{url}
\usepackage{hyperref}
\definecolor{myblue}{RGB}{0, 102, 204} 
\hypersetup{
    colorlinks=true,        
    linkcolor=myblue,         
    citecolor=myblue,          
    urlcolor=magenta,       
    filecolor=cyan          
}
\usepackage{mathtools}
\usepackage{wrapfig}
\usepackage{stfloats}
\def\BibTeX{{\rm B\kern-.05em{\sc i\kern-.025em b}\kern-.08em
    T\kern-.1667em\lower.7ex\hbox{E}\kern-.125emX}}
\begin{document}

\title{
VectorPainter: Advanced Stylized Vector Graphics Synthesis Using Stroke-Style Priors
}

\author{
    Juncheng Hu, Ximing Xing, Jing Zhang, Qian Yu\textsuperscript{$\dagger$} \\
    Beihang University \\
    {\tt\small \{hujuncheng, ximingxing, zhang\_jing, qianyu\}@buaa.edu.cn} \\
    \small{\textsuperscript{$\dagger$}Corresponding Author.}
}

\twocolumn[{
\renewcommand\twocolumn[1][]{#1}
\maketitle
\begin{center}
\captionsetup{type=figure}
\vspace{-1em}
\includegraphics[width=\textwidth]{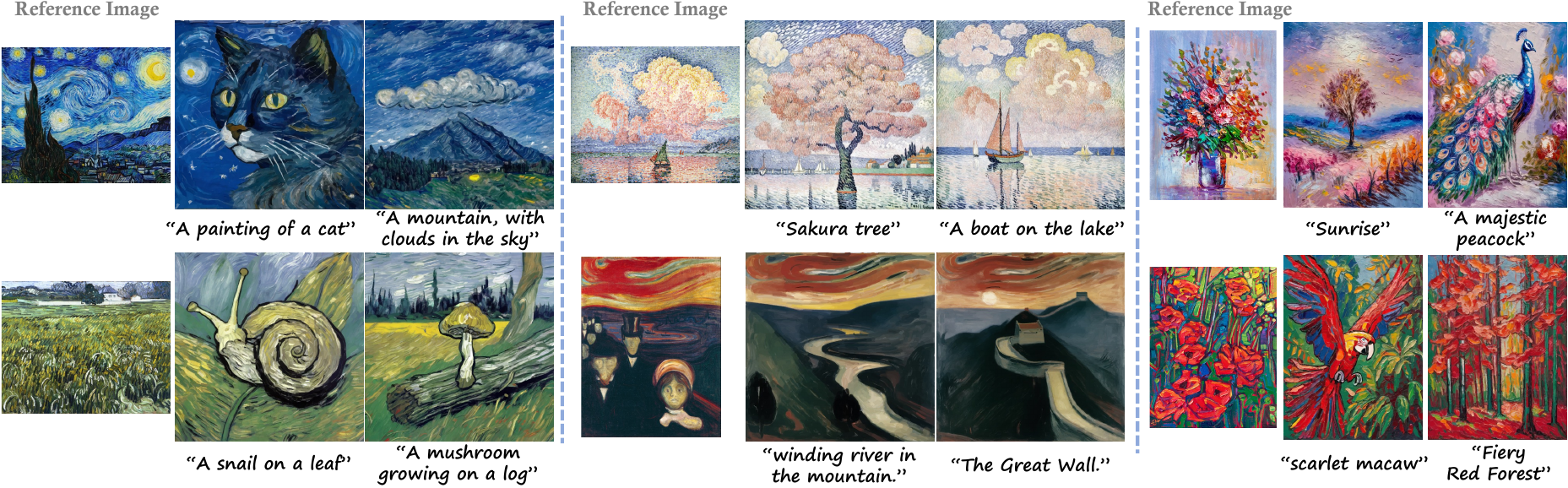}
\captionof{figure}{
\textbf{Vector Graphics Synthesized by Our VectorPainter.}
Given a bitmap reference image and a textual description, our VectorPainter generates a stylized vector graphic aligned with the text.}
\vspace{-0.6em}
\label{fig:teaser}
\end{center}
}]


\begin{abstract}
We introduce \textit{VectorPainter}, a novel framework designed for reference-guided text-to-vector-graphics synthesis. Based on our observation that the style of strokes can be an important aspect to distinguish different artists, our method reforms the task into synthesizing a desired vector graphic by rearranging stylized strokes, which are vectorized from the reference images. Specifically, our method first converts the pixels of the reference image into a series of vector strokes, and then generates a vector graphic based on the input text description by optimizing the positions and colors of these vector strokes. To precisely capture the style of the reference image in the vectorized strokes, we propose an innovative vectorization method that employs an imitation learning strategy. To preserve the style of the strokes throughout the generation process, we introduce a style-preserving loss function. Extensive experiments have been conducted to demonstrate the superiority of our approach over existing works in stylized vector graphics synthesis, as well as the effectiveness of the various components of our method. Code, model, and data will be released at: \href{https://hjc-owo.github.io/VectorPainterProject/}{https://hjc-owo.github.io/VectorPainterProject/}
\end{abstract}

\section{Introduction}
\noindent In recent years, there has been a growing number of studies focused on vector graphics or SVG (Scalable Vector Graphics) synthesis due to their superior compatibility with visual design applications compared to raster images~\cite{ma2022towards, frans2022clipdraw, schaldenbrand2022styleclipdraw,vinker2022clipasso,vinker2023clipascene,jain2023vectorfusion, xing2024diffsketcher, xing2023svgdreamer,xing2025svgdreamer++,xing2025empowering,xing2024svgfusion}. Particularly driven by success in text-to-image (T2I) models~\cite{rombach2022high, podell2023sdxl}, recent works such as VectorFusion~\cite{jain2023vectorfusion} and SVGDreamer~\cite{xing2023svgdreamer} have explored the task of text-to-SVGs synthesis by integrating T2I models with a differentiable rasterizer~\cite{li2020differentiable}. Despite the advancements, precisely controlling the style of generated vector graphics using only text descriptions remains challenging.

Reference images provide an effective mechanism for precise style control in synthesized images~\cite{kolkin2019style, wang2024instantstyle, hertz2024style,chung2024style}. However, the task of reference-guided text-to-SVGs remains underexplored. 
Existing methods like StyleCLIPDraw~\cite{schaldenbrand2022styleclipdraw} and NST-VG~\cite{efimova2023neural} integrate style transfer approaches designed for raster images into the optimization-based text-to-SVG synthesis methods. These methods begin by rasterizing randomly initialized vector graphics using differentiable rasterizers, then compute style and content losses.
Unfortunately, these methods struggle to produce high-quality stylized vector graphics, as the gradients computed in the pixel space do not effectively guide the optimization of vector primitives.

\begin{figure}[t]
\centering
\includegraphics[width=1.0\linewidth]{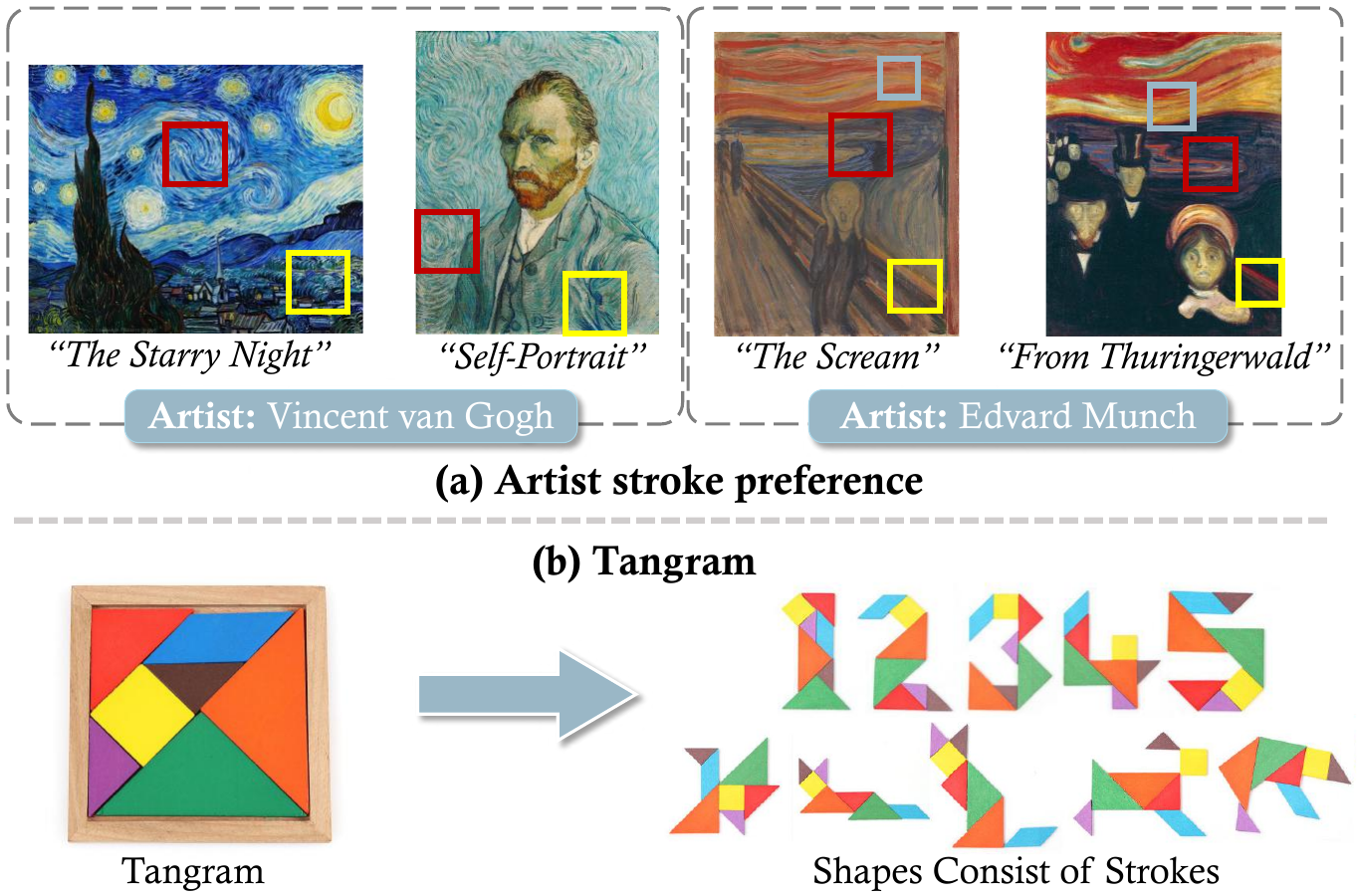}
\caption{
\textbf{Motivation of Our Method.} \textbf{(a)} Artists exhibit distinct stroke preferences; patches highlighted in the same color share a similar stroke style.
\textbf{(b)} A Tangram puzzle forms various objects with basic elements, inspiring us to treat stylization as a rearrangement of strokes from the reference image.
}\label{img: motivation}
\vspace{-1.5em}
\end{figure}
The key idea of this work is leveraging the stroke style of the reference image for stylized vector graphics synthesis. As illustrated in Fig.~\ref{img: motivation}(a), we observe that different artists can be distinguished by the style of strokes in their paintings, such as width, colors, opacity, and arrangement of the strokes. Therefore, one can imitate the style of the artist by learning their preference for stroke usage. 
A vector graphic is composed of a series of primitives, such as Bézier curves, analogous to how a painting is composed of a series of strokes. In other words, a stroke in a painting can be modeled as a primitive in a vector graphic. 
Furthermore, inspired by the Tangram puzzle which forms different objects using a set of basic elements, as illustrated in Fig.~\ref{img: motivation}(b), we reform the synthesis task as a rearrangement of a set of strokes extracted from the reference image.

We propose a novel framework, named \textbf{VectorPainter}, for generating stylized vector graphics. Given a text prompt and a style reference image, VectorPainter is capable of producing a vector graphic whose content aligns with the text prompt while the style remains faithful to the reference image. 
Our model comprises two main components: Stroke Style Extraction and Stylized SVG Synthesis. In the \textit{stroke style extraction stage}, we introduce a new method to extract a set of vectorized strokes from the reference image. Drawing inspiration from Berger \textit{et al.}~\cite{berger2013style} and recent advancements in stroke-based rendering~\cite{zou2021stylized, kotovenko2021rethinking, hu2024supersvg}, our technique identifies features including local colors and the directionality of strokes to facilitate stroke extraction. 
We further introduce a learning strategy to ensure the extracted strokes precisely capture the style of the reference image. 
In the \textit{SVG synthesis stage}, we follow prior works~\cite{jain2023vectorfusion,xing2024diffsketcher,xing2023svgdreamer,xing2025svgdreamer++} to adopt an optimization-based pipeline by combining a T2I model, \textit{i.e.}, LDM~\cite{rombach2022high}, and a differentiable rasterizer~\cite{li2020differentiable}.
VectorPainter utilizes the extracted strokes to initialize the target vector graphic, setting a good starting point for subsequent optimization. 
During optimization, these strokes are rearranged, and their colors are tuned, to create new content as specified by the text prompt, entirely within the vector space.
Furthermore, to preserve the style of the extracted strokes throughout the synthesis process, we introduce a style-preserving loss, including an optimal transportation loss and a DDIM inversion~\cite{song2021ddim} method. 
Through harnessing the stroke style of the reference image, our VectorPainter can synthesize high-quality vector graphics as desired, as shown in Fig.~\ref{fig:teaser}.

\noindent In summary, the contributions of this work are threefold:
\begin{itemize}[left=0pt]
\item We introduce a new method, named VectorPainter, which innovatively conceptualizes the task of stylized vector graphics generation as a process of rearranging the strokes extracted from the reference image. 
\item We propose a novel algorithm for extracting a set of vectorized strokes from the reference image. These strokes serve as the basic elements for forming the new content. Additionally, we introduce a style-preserving loss to maintain the style of stroke throughout the generation process.
\item We conduct extensive experiments to assess the effectiveness of our model and individual components. 
The results demonstrate the superiority of VectorPainter in producing high-quality stylized SVGs.
\end{itemize}

\section{Related Work}
\subsection{Vector Graphics Synthesis}
\noindent Scalable Vector Graphics (SVGs) are comprised of essential components such as B\'ezier curves, lines, shapes, and colors to represent images. The latest technique for generating SVGs involves using a differentiable rasterizer such as DiffVG~\cite{li2020differentiable}. DiffVG bridges the gap between vector graphics and raster image spaces, allowing for the generation of vector graphics without requiring access to traditional vector graphics datasets like those used in earlier vector graphics synthesis methods. Numerous studies~\cite{frans2022clipdraw,schaldenbrand2022styleclipdraw,vinker2022clipasso,vinker2023clipascene} use the CLIP model~\cite{radford2021learning} to supervise the synthesis of SVGs, while others~\cite{jain2023vectorfusion,xing2024diffsketcher,xing2023svgdreamer,xing2025svgdreamer++} combine the text-to-image (T2I) diffusion model~\cite{rombach2022high} with a differentiable rasterizer~\cite{li2020differentiable} for SVG synthesis. Both of them achieved impressive results.

\subsection{Style Transfer}
\noindent Style Transfer is a task in computer vision that involves combining a content image and a style image to create a new image that preserves the former's content and the latter's style patterns. Over the years, many researchers have proposed various models~\cite{kolkin2019style, wang2024instantstyle, hertz2024style,chung2024style} to improve the quality and speed of style transfer. However, research has mainly focused on raster images, with only limited work on vector graphics.

StyleCLIPDraw~\cite{schaldenbrand2022styleclipdraw} attempted to enhance CLIPDraw~\cite{frans2022clipdraw} by incorporating a style loss to achieve stylized vector graphics synthesis.
However, the results were often messy and the style representation was inadequate. 
Efimova \textit{et al.}~\cite{efimova2023neural} attempted to transfer the style of an SVG onto another. However, this method not only produced undesired outputs but also proved impractical for most applications, as it requires that both the content and style controls be vector images.

\section{Methodology}
\noindent In this section, we introduce \textit{VectorPainter} for stylized vector graphics synthesis. 
Given a text prompt $\mathcal{P}$ and a reference painting image $\mathcal{I}_{s}$, VectorPainter aims to generate a vector graphic $\mathcal{S}$ whose content aligns with the text prompt while the style remains consistent with the reference image. 
VectorPainter comprises two main steps, \textit{Stroke Style Extraction} and \textit{Stylized SVG Synthesis}. In the stroke style extraction stage, VectorPainter extracts vectorized style strokes from the reference image. During the Stylized SVG Synthesis stage, VectorPainter produces SVGs based on the style strokes extracted in the first step. A style-preserving loss is introduced to enhance the stylistic consistency of the final output SVGs.

\begin{figure}
\centering
\includegraphics[width=\linewidth]{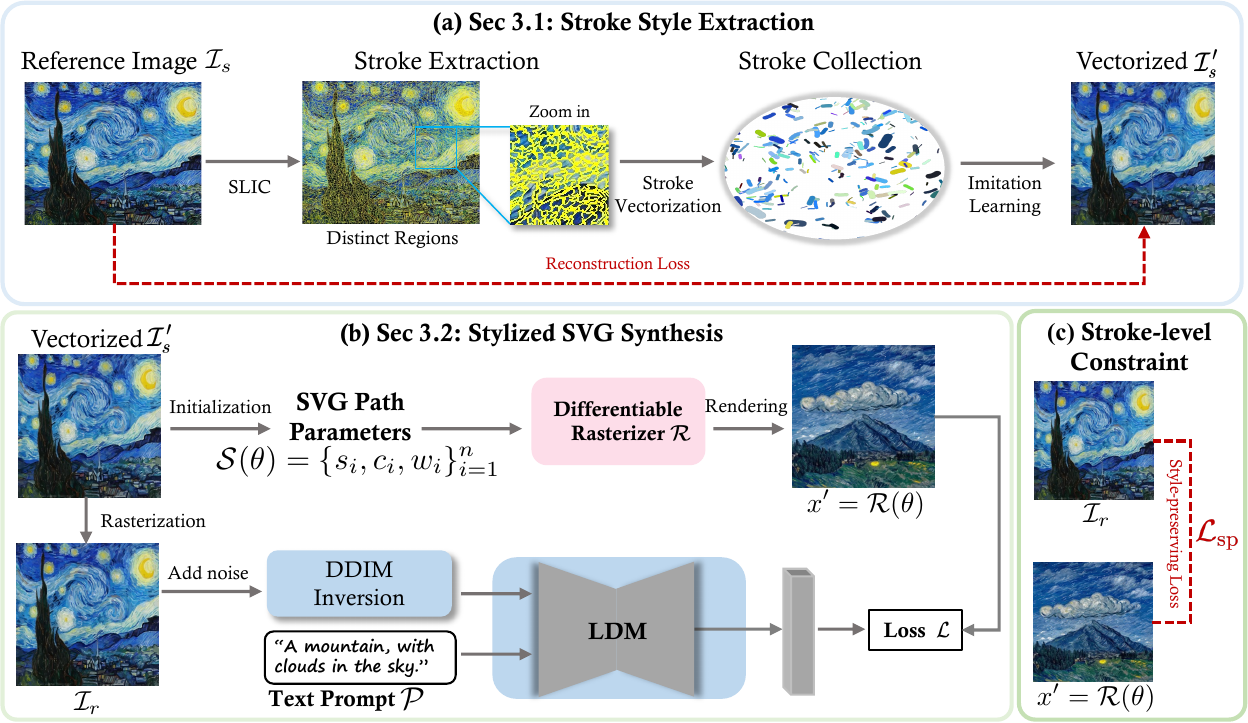}
\vspace{-1.5em}
\caption{
\textbf{Pipeline of Our VectorPainter.} 
Given a text prompt $\mathcal{P}$ and a reference image $\mathcal{I}_{s}$, VectorPainter optimizes the parameters $\theta$ of the vector graphic $\mathcal{S}$.
Initially, it extracts vectorized strokes from $\mathcal{I}_s$, which are then used to initialize the synthesis process. During synthesis, a style-preserving loss is introduced to ensure the style fidelity.
}\label{img:pipeline}
\vspace{-1.5em}
\end{figure}
\subsection{Stroke Style Extraction}
\label{subsec:stroke_extract}
\subsubsection{Stroke Extraction}
We define the desired vector graphic $\mathcal{S}$ as a collection of $n$ vector strokes $\mathcal{S} = \{E_i\}_{i=1}^n$.
Specifically, each stroke $E_i = \{s_i, c_i, w_i\}$ is represented by a quadratic B\'ezier curve, defined by three control points, denoted as $P_i=\{p_i^j\}_{j=1}^3=\{(x_i, y_i)^j\}_{j=1}^3$, where $x$ and $y$ represent the coordinates within the canvas. 
The stroke color and width are defined as $c_i=\{r,g,b,a\}$ and $w_i$, respectively.

Although painting an image is a dynamic process, the resultant style reference image is static, composed of unordered pixels. Consequently, it is challenging to directly extract the original strokes from the reference image.
However, we observed that most strokes can be differentiated from one another, as the pixels belonging to the same stroke exhibit similar attributes, such as color and texture. 
Based on this observation, we employ a superpixel method to extract strokes.
As depicted in Fig.~\ref{img: recons}(a), given a style image, we use SLIC~\cite{Achanta2010SLIC} to partition pixels exhibiting similar features within the image into distinct regions. Each region is treated as a style stroke in the reference image and is subsequently vectorized.

\subsubsection{Vectorized Stroke Imitation Learning}
\label{sec:imitation_learning}
\begin{figure}[t]
\centering
\includegraphics[width=0.9\linewidth]{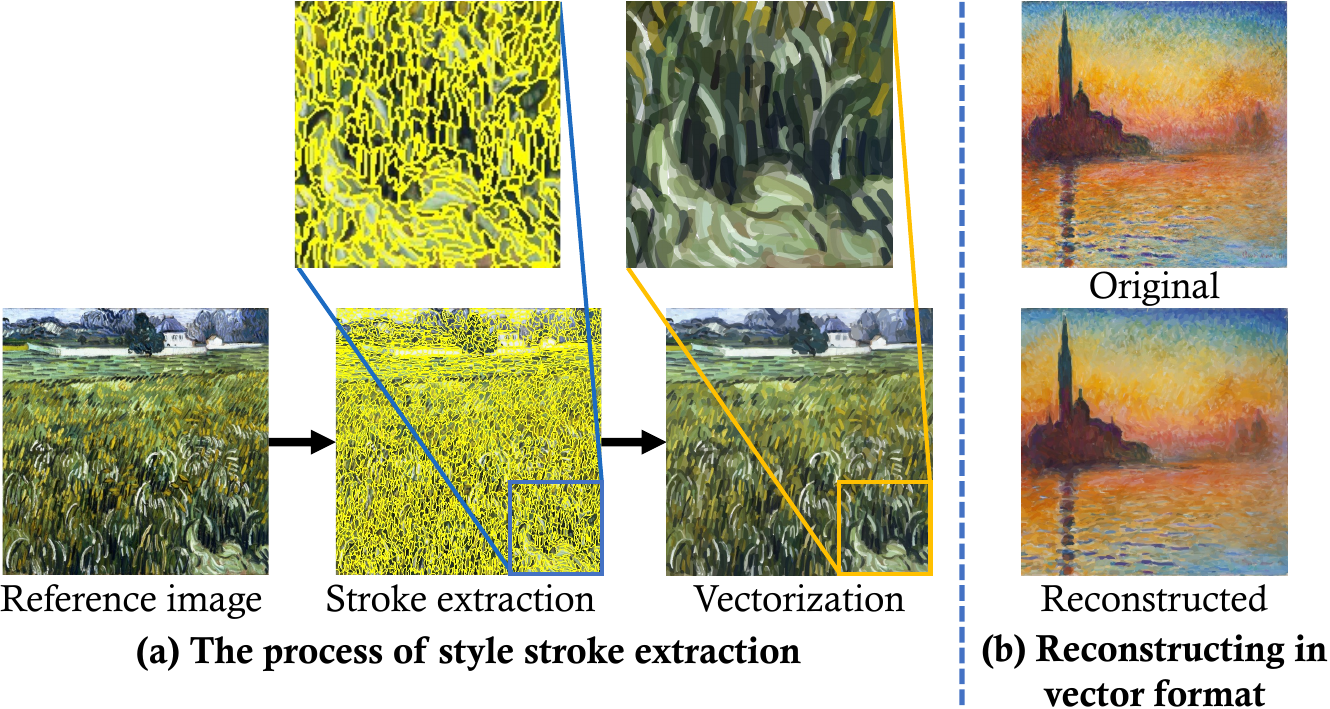}
\caption{
\textbf{Pipeline of Stroke Style Extraction.}
(a) The process comprises two main steps: stroke extraction and stroke vectorization.
(b) VectorPainter accurately reconstructs the reference image using the extracted vector strokes.
} \label{img: recons}
\vspace{-1.5em}
\end{figure}

We propose that stylized vector drawings should begin by generating strokes that closely resemble the reference image within the vector domain. This approach, referred to as \textit{vector stroke imitation learning}, ensures better style consistency and structural alignment.
To extract a vector stroke from a segmented region, we identify the pair of points with the maximum mutual distance within this region as the initial and terminal control points. The stroke color is determined by averaging the colors within the region, while the stroke thickness is calculated as the average distance between the border points and the central line, which connects the control points.
Specifically, once we obtain the vectorized strokes, we produce a vectorized version of the reference image, denoted as $\mathcal{I}_{s}'$. 
We then rasterize $\mathcal{I}_{s}'$ and compare it with $\mathcal{I}_{s}$ using a mean-square-error loss. 
$\mathcal{I}_{s}'$ undergoes optimization over $N_{i}$ iterations, during which the quality of the vectorized strokes can be further improved, as illustrated in Fig.~\ref{img: recons}(b).

\subsection{Stylized SVG Synthesis}
\label{subsec:svg_rendering}
\noindent In this step, we adopt an optimization-based pipeline for SVG synthesis, following the prior work~\cite{jain2023vectorfusion,xing2024diffsketcher,xing2023svgdreamer}.
Initially, an SVG is created in the vector space and subsequently rendered to the pixel space using DiffVG~\cite{li2020differentiable}. In the pixel space, losses are computed, then gradients are backpropagated into the vector space to optimize the parameters of the SVG strokes. 
To ensure that the output SVG exhibits the desired style, we propose initializing the SVG with the style strokes extracted from the reference image, complemented by a style-preserving loss. 

\noindent\textbf{1) SVG Initialization with Style Strokes.}
As depicted in Fig.~\ref{img: recons}(b), the style strokes enable the accurate reconstruction of the original reference image, effectively preserving its intricate details. These vectorized strokes encapsulate the style of the reference image and can provide style priors for SVG generation. Therefore, we propose using these style strokes to initialize the SVG for subsequent optimization. Compared to random initialization, this approach significantly reduces the difficulty in synthesizing SVGs that accurately reflect the target style.

\noindent\textbf{2) Style-Preserving Loss.}
To preserve style consistency during the optimization process and minimize changes to individual strokes, we introduce a style-preserving loss.
The style-preserving loss comprises two components: one involves stroke-level constraints implemented through optimal transport loss, while the other pertains to global-level style constraints determined by the perceived similarity between the rendering of the generated SVG and the reference image. 

\noindent\textit{Stroke-level constraint.} At the stroke level, we implement constraints through optimal transport, specifically using the Sinkhorn distance~\cite{cuturi2013sinkhorn,luise2018differential}. 
This optimal transport loss $\mathcal{L}_{\mathrm{ot}}$ measures the minimum effort required to move strokes from one location to another, thereby discouraging excessive variation in the synthesized vector graphics.

\noindent\textit{Global-level constraint.} For global consistency, we incorporate DDIM inversion~\cite{song2021ddim}  to maintain the overall style between the output SVG and the reference image. Initially, the reference image is encoded into a latent space. Subsequently, using the DDIM scheduler, noise is introduced to this latent representation. This noisy latent representation then serves as input to LDM~\cite{rombach2022high,podell2023sdxl}, enabling the generation of images that closely mirror the target style.


\section{Experiments}
\noindent In this section, we compare our VectorPainter with a series of baseline methods and perform ablation studies to evaluate the contribution of each component in our approach.

\begin{figure*}
\centering
\includegraphics[width=\textwidth]{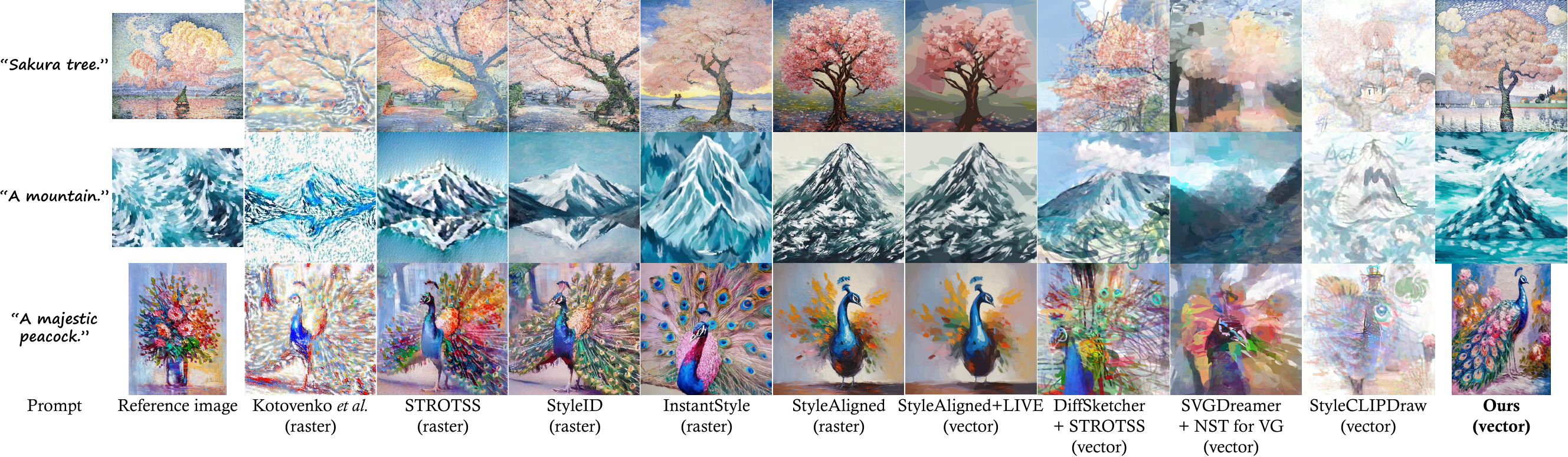}
\vspace{-1.75em}
\caption{
\textbf{Qualitative Comparison with Baseline Methods.}
The terms ``vector'' and ``raster'' in parentheses indicate the format of the generated results, where ``vector'' refers to vector graphics and ``raster'' refers to raster graphics. The image style transfer algorithm effectively preserves style consistency, as shown in the fourth to sixth columns. 
However, it disrupts the structure of vector-based results due to the inconsistency between the style loss and the vector content synthesis loss during optimization, as shown in the seventh to tenth columns.
} \label{img:compare}
\vspace{-1.7em}
\end{figure*}
\subsection{Evaluation Metrics}
\noindent To demonstrate the effectiveness of our method, we employ four metrics to quantitatively evaluate our approach: \textit{(1) CLIP Score}~\cite{radford2021learning}. This metric measures the alignment of the synthesized SVGs with the text prompts, assessing how well the generated graphics match the text description. \textit{(2) LPIPS}~\cite{zhang2018unreasonable}. This metric quantifies the content fidelity between the stylized image and its corresponding content image, indicating how accurately the content is preserved in the stylization process. 
\textit{(3) FID}~\cite{heusel2017gans}. This metric measures the style fidelity between the stylized image and its respective style reference, evaluating how closely the synthesized image matches the style of the reference image.
\textit{(4) ArtFID}~\cite{wright2022artfid}. Designed to evaluate the preservation of both content and style, ArtFID is acknowledged for its strong alignment with human judgment. 
The ArtFID is computed as ArtFID = (1 + LPIPS) · (1 + FID).

\subsection{Comparison Baselines}
\noindent To synthesize stylized vector graphics, there are mainly three approaches: \textit{(1) Synthesis though Text Prompt and Reference Image}. Like the existing method StyleCLIPDraw~\cite{schaldenbrand2022styleclipdraw} and our VectorPainter,  these methods generate stylized vector graphics directly based on a given text and a reference image. 
\textit{(2) Rasterization then Vectorization}. This approach involves performing style transfer in the pixel space and then converting the raster image into the vector space. For example, ``StyleID~\cite{chung2024style} + LIVE~\cite{ma2022towards}''. 
\textit{(3) Optimization-based Methods with Style Transfer Supervision}. This approach combines an SVG synthesis method with a style transfer method. 
For instance, the baseline method ``DiffSketcher~\cite{xing2024diffsketcher} + STROTSS~\cite{kolkin2019style}'' incorporates the style transfer method STROTSS into DiffSketcher as part of the supervision for optimizing an SVG. 

Another benchmark is SVGDreamer~\cite{xing2023svgdreamer}, which controls style through text and primitive types. However, it does not support style control via reference images. To address this limitation, we incorporate the style loss from~\cite{efimova2023neural} into its optimization process, enabling reference image-based style transfer. 
We refer to this modified version as ``SVGDreamer~\cite{xing2023svgdreamer} + NST for VG~\cite{efimova2023neural}'', which enforces style transfer supervision directly in the vector space.

Additionally, we include several SOTA style transfer methods designed for raster images for comparison, including STROTSS, StyleID, InstantStyle~\cite{wang2024instantstyle}, and StyleAligned~\cite{hertz2024style}, as well as a stroke-based rendering method Kotovenko \textit{et al.}~\cite{kotovenko2021rethinking}. Note that the outputs of these methods are raster images.

\subsection{Qualitative Evaluation}
\noindent From 
Fig.~\ref{img:compare}, we can make the following observations: \textbf{(1)} The results of StyleCLIPDraw are poor, indicating the challenges of directly synthesizing stylized vector graphics. 
\textbf{(2)} Most raster-image-oriented style transfer methods preserve the reference's style well. 
However, their results are in raster format.
Once vectorization is performed, the quality of the resultant SVGs significantly decreases due to the lossy nature of vectorization (``StyleAligned'' vs. ``StyleAligned + LIVE''). 
\textbf{(3)} ``DiffSketcher + STROTSS'' performs relatively well but fails to adequately preserve the reference style. ``SVGDreamer + NST for VG'' performs even worse, indicating that simply integrating style transfer methods with SVG synthesis techniques does not necessarily lead to satisfactory results.
\textbf{(4)} In comparison, our VectorPainter performs the best, producing high-quality SVGs whose content aligns with the input text and whose style remains consistent with the reference image.

\begin{figure}
\centering
\includegraphics[width=0.9\linewidth]{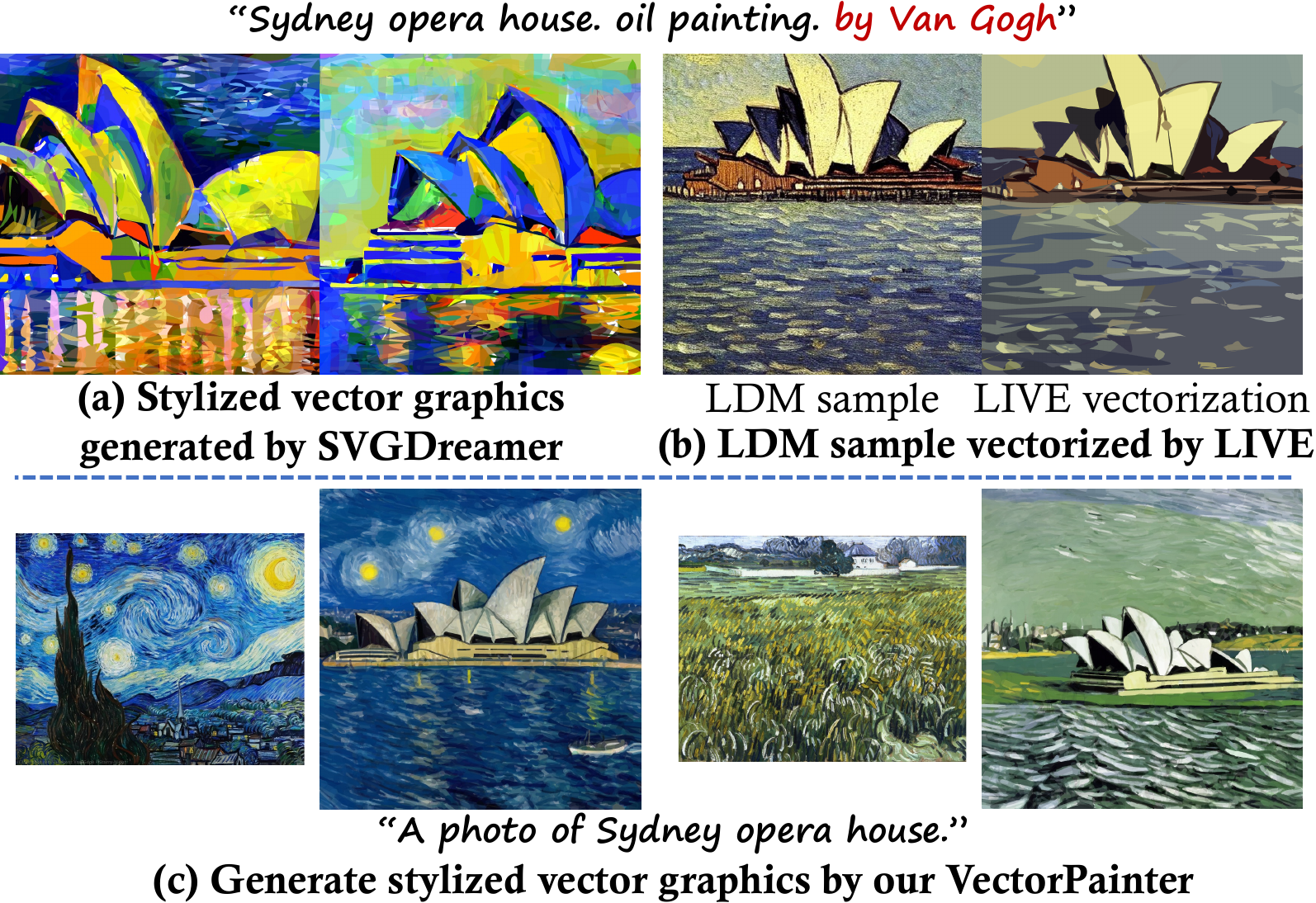}
\vspace{-0.35em}
\caption{
\textbf{Comparison of Various Stylized Control Methods.}
(a) SVGDreamer~\cite{xing2023svgdreamer} utilizes text and primitive types to control the style of vector graphics.
(b) Rasterization followed by Vectorization: An image is first generated using a text-to-image method~\cite{rombach2022high} and then vectorized via LIVE~\cite{ma2022towards}.
(c) Ours (VectorPainter): A more advanced vector graphics stylization method that supports both reference images for style control and text descriptions for content control.
} \label{img: control_by_prompt}
\vspace{-0.5em}
\end{figure}
\begin{table}
\centering
\caption{
\textbf{Quantitative Evaluation.}
The format of the generated results is indicated in parentheses.
}
\label{table: comparison}
\resizebox{0.9\linewidth}{!}{
\begin{threeparttable}
\begin{tabular}{c|cccc}
\toprule
& ArtFID$\downarrow$ & FID$\downarrow$ & LPIPS$\downarrow$~\tnote{1} & CLIPScore$\uparrow$ \\
\midrule
Kotovenko \textit{et al.} (raster) & 61.331 & 33.406 & 0.783 & 0.3043 \\
STROTSS (raster) & 42.196 & 23.985 & 0.689 & 0.2845 \\
StyleID (raster) & 37.724 & 23.884 & 0.516 & 0.2930 \\
InstantStyle (raster)~\tnote{2} & - & 23.015 & - & 0.2628 \\
StyleAligned (raster)~\tnote{2} & - & \textbf{22.538} & - & 0.2576 \\
\midrule
StyleAligned~\tnote{2} + LIVE (vector) & - & 36.474 & - & 0.2925 \\
DiffSketcher + STROSS (vector) & 49.414 & 30.562 & 0.566 & 0.2817 \\
SVGDreamer + VectorNST (vector) & 80.335 & 44.869 & 0.751 & 0.2506 \\
StyleCLIPDraw (vector)~\tnote{3} & - & 32.572 & - & 0.2831 \\
\textbf{Ours (vector)} & \textbf{26.962} & 23.160 & \textbf{0.116} & \textbf{0.3109} \\
\bottomrule
\end{tabular}
\begin{tablenotes}
\item[1] If a text-guided synthesis method involves using the LDM samples, we employ the LDM sample as the content image.
\item[2] InstantStyle and StyleAligned incorporate some modifications to the LDM. As they generate stylized images directly without relying on content images, we do not calculate their LPIPS scores.
\item[3] Since StyleCLIPDraw does not involve content image during its generation process, we do not calculate its LPIPS score.
\end{tablenotes}
\vspace{-2em}
\end{threeparttable}
}
\end{table}
To better illustrate the advantages of incorporating a reference image for style control, we conduct a comparative analysis between our approach and SVGDreamer~\cite{xing2023svgdreamer}, which relies solely on text prompts for synthesizing stylized vector graphics. 
As illustrated in Fig.~\ref{img: control_by_prompt} (a), employing only text prompts fails to achieve precise control over style. 
Additionally, we present results from vectorizing a sample produced by the LDM in Fig.~\ref{img: control_by_prompt} (b).  
While both methods successfully generate correct content, the style does not correspond with that specified by the text prompts.
In comparison, using a reference image can realize more precise control, thus our results are better, as depicted in Fig.~\ref{img: control_by_prompt} (c).

\subsection{Quantitative Evaluation}
\label{sec:quantitative}
\noindent Table~\ref{table: comparison} presents the quantitative evaluation of different methods. Our approach outperforms all baselines across most metrics, demonstrating superior style transfer performance in the vector domain. Specifically, our method achieves the lowest ArtFID (26.962), indicating a high degree of artistic style fidelity, and the lowest LPIPS (0.116), suggesting the best perceptual similarity to the reference style. Additionally, our method achieves the highest CLIPScore (0.3109), reflecting strong alignment with the input content and style descriptions.

Among the raster-based methods, both StyleAligned (FID 22.538) and StyleID (LPIPS 0.516) achieve high-quality stylization. However, these methods operate in the raster domain and are not directly comparable to vector-based approaches.

For vector-based methods, DiffSketcher + STROTSS performs better than SVGDreamer + VectorNST, achieving lower ArtFID (49.414 vs. 80.335) and FID (30.562 vs. 44.869). However, both methods still fall short in terms of style preservation and content fidelity compared to our approach. StyleCLIPDraw, while achieving a reasonably low FID (32.572), does not provide LPIPS scores due to its generation process.

Overall, VectorPainter outperforms other vector-graphics-oriented baseline methods and even surpasses raster-image-oriented methods in terms of ArtFID, LPIPS, and CLIP scores. Our approach achieves significant improvements in both artistic style fidelity and content preservation, establishing a new benchmark for style transfer in the vector domain. These results suggest that VectorPainter effectively balances style fidelity and content preservation.

\subsection{Ablation Studies}
\label{sec:abs}
\begin{figure}[t]
\centering
\includegraphics[width=0.85\linewidth]{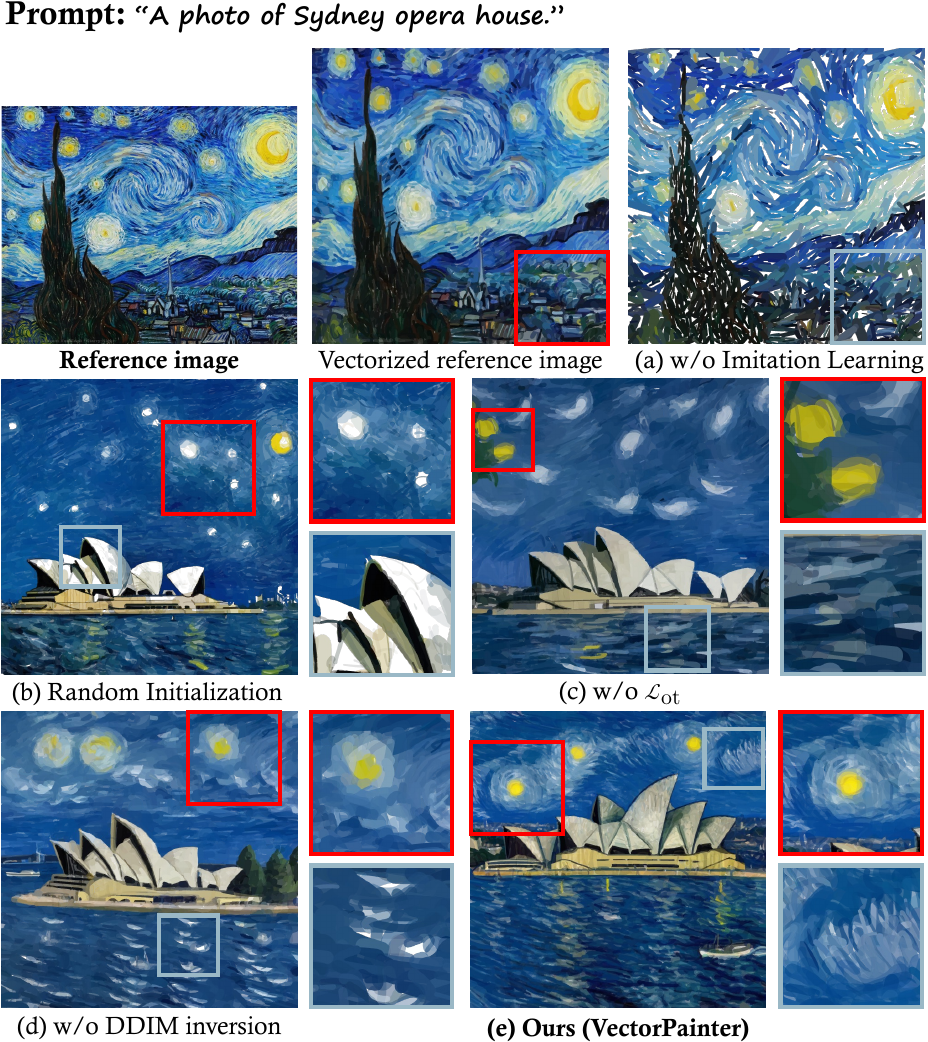}
\vspace{-0.35em}
\caption{
\textbf{Effect of Different Components of VectorPainter.}
\textbf{(a)} Comparison of simulation quality with and without imitation learning (as described in Sec.~\ref{sec:imitation_learning}).
\textbf{(b)} Evaluation of our stroke initialization method versus random initialization.
\textbf{(c)} Assessment of the effectiveness of Optimal Transport Loss $\mathcal{L}_\mathrm{ot}$ (as described in Sec.~\ref{subsec:svg_rendering}).
\textbf{(d)} The impact of DDIM inversion.
\textbf{(e)} Performance of the full VectorPainter model.
} \label{img: Ablation}
\vspace{-1.75em}
\end{figure}
\noindent\textbf{1) Effect of Imitation Learning Strategy.}
Our imitation learning strategy aims to ensure that strokes extracted from the reference image authentically capture the desired style. As shown in Fig.~\ref{img: Ablation}(a), without this strategy, the extracted strokes inadequately reflect the reference style, resulting in noticeable blank holes in the reconstructed image. This highlights the importance of imitation learning for accurate style replication in vector graphics. 

\noindent\textbf{2) Effect of Our Initialization Strategy.}
In VectorPainter, we propose using strokes extracted from the reference image to initialize the target vector graphics, providing an advantageous starting point for subsequent optimization. 
To demonstrate its effectiveness, we conducted comparisons with random initialization. 
As depicted in Fig.~\ref{img: Ablation}(b), results from random initialization exhibit incomplete regions and disorganized strokes, like the stars and the Sydney Opera House.

\noindent\textbf{3) Effect of Style-Preserving Loss.}
From Fig.~\ref{img: Ablation}(c) and (d), we can have the following observations:
\textit{(1) Without Optimal Transport Loss $\mathcal{L}_\mathrm{ot}$}: The strokes in the final SVG significantly deviate from those in the reference image, resulting in a loss of style fidelity.
\textit{(2) Without DDIM inversion}: Using standard LDM leads to smoother color distributions and less defined strokes. When using DDIM inversion, as indicated in Fig.~\ref{img: Ablation}(e), the result can better capture and replicate the style of the reference image.

\section{Conclusion \& Discussion}
\noindent In this work, we introduce \textit{VectorPainter}, a novel and effective approach for synthesizing stylized vector graphics using text prompts and reference images. VectorPainter posits that the style of strokes uniquely characterizes the overall style of a painting. This work is the first to conceptualize the stylization process as the re-organization of vectorized strokes extracted from the reference image. Comprehensive experimental results validate the effectiveness of each component within our proposed model.


\section*{Acknowledgment}
This work was supported in part by the Young Elite Scientists Sponsorship Program by the Chinese Association for Science and Technology (CAST), in part by Huawei-BUAA Joint Lab, in part by the Fundamental Research Funds for the Central Universities, in part by National Natural Science Foundation of China (No.62461160331, No.62132001), and in part by the Beihang World TOP University Collaboration Program.

{
    \small
    \bibliographystyle{IEEEbib}
    \bibliography{references}
}

\clearpage
\renewcommand{\thefigure}{S\arabic{figure}}
\setcounter{figure}{0}
\renewcommand{\thetable}{S\arabic{table}}
\setcounter{table}{0}
\renewcommand{\thealgorithm}{S\arabic{algorithm}}
\setcounter{algorithm}{0}
\renewcommand{\theequation}{S\arabic{equation}}
\setcounter{equation}{0}


\twocolumn[
    \begin{center}
        {\LARGE Supplementary Material}
    \end{center}
]

\section*{Overview}
\label{sec:overview}
In this supplementary material, we provide additional details and discussions related to our work on \textbf{VectorPainter}. Specifically, it covers the following aspects:
\begin{itemize}
\item In Section~\ref{supp:loss_func}, we provide the detail of the total loss function used in Sec.~\ref{subsec:svg_rendering}.
\item In Section~\ref{supp:implementation}, we explain the implementation details of our VectorPainter.
\item In Section~\ref{supp:stroke_extraction}, we show the algorithm of our Stroke Style Extraction Algorithm.
\item In Section~\ref{supp:more}, we present additional qualitative results of VectorPainter, demonstrating its capability to generate stylized SVGs with high visual quality.
\item In Section~\ref{supp:user_study}, we conducted a user study to compare our method with the baseline methods.
\item In Section~\ref{supp:limitations}, we explain some limitations of our method.
\end{itemize}

\begin{algorithm*}
\caption{Stroke Extraction Algorithm}
\begin{algorithmic}[1]
\Require{Reference image $\mathcal{I}_s$ and distinct regions $T$ of segmented reference image $\mathcal{I}_s'$}
\Ensure{List of initialized strokes $\mathcal{S} = \{E_i\}_{i=1}^n = \{s_i, c_i, w_i\}_{i=1}^n$ }
\algnewcommand\algorithmicforeach{\textbf{for each}}
\algdef{S}[FOR]{ForEach}[1]{\algorithmicforeach\ #1\ \algorithmicdo}
\State\textbf{Initialize:} An empty parameters set $\mathcal{S}$
\ForEach{$region \in T$} \Comment{$n$ regions in $T$}
\State Calculate the border points $border$ of $region$
\State Calculate distance $dist$ of each point pair
\State Find the point pair $(p_1, p_3)$ where $dist(p_1, p_3) = \max(dist)$
\State $p_2\gets\dfrac{p_1+p_3}{2}$
\State Stroke control points $s_i\gets\{p_1, p_2, p_3\}$
\State Stroke width $w_i \gets \dfrac{1}{M} \sum_{j=1}^{M} \| border[j], \overrightarrow{p_1p_3} \|$, where $M$ indicates the number of border points
\State Stroke color $c_i \gets \dfrac{1}{N} \sum_{k=1}^{N} region.point[k].color$, where $N$ represents the number of points in $region$
\State $\mathcal{S}.append(\{s_i, c_i, w_i\})$
\EndFor \Comment{Here, $\mathcal{S}$ represents the parameters list of the vectorized version of the reference image $\mathcal{I}_s'$}
\While{not converged} \Comment{Set to $N_{i}$ iteration steps}
\State Render the SVG parameter $\theta$ to get a raster image $x = \mathcal{R}(\theta)$.
\State $\mathcal{S}\gets\mathcal{S} - \eta\|x - \mathcal{I}_s\|_2$
\EndWhile
\State\Return $\mathcal{S}$
\end{algorithmic}
\label{algo:supp_stroke_extraction}
\end{algorithm*}

\section{Total Loss Function}
\label{supp:loss_func}
\noindent\textbf{Optimal Transport Loss $\mathcal{L}_\mathrm{ot}$.} \quad
Unlike pixel-based losses, such as the $\ell_2$ loss that computes the mean squared error (MSE) for each corresponding pixel, the minimum transportation loss provides a more effective measure of similarity between the canvas and the reference image.

Similar to neural painting~\cite{zou2021stylized}, for a rendered canvas $x'=\mathcal{R}(\theta)$ and rendered version of $\mathcal{I}_s'$, denoted as $\mathcal{I}_r$, we use a smoothed variant of the classic optimal transport distance~\cite{luise2018differential}, enhanced with an entropic regularization term, to measure the similarity between $x'$ and $\mathcal{I}_r$. This approach results in the well-known Sinkhorn distance~\cite{cuturi2013sinkhorn}. Let $p_{x'}$ and $p_r$ as their probabilistic marginal functions, respectively. We define $\mathbf{P} \in \mathbb{R}_+^{n \times n}$ as the joint probability matrix, where the $(i, j)$-th element represents the joint probability of the $i$-th pixel in $x'$ and the $j$-th pixel in $\mathcal{I}_r$. Here, $n$ is the total number of pixels in the rendered image. Similarly, we define $\mathbf{D}$ as the cost matrix, where the $(i, j)$-th element represents the Euclidean distance between the location of the $i$-th pixel in $x$ and the $j$-th pixel in $\mathcal{I}_r$. Thus, $\mathbf{D}$ encodes the ``labor cost'' of transporting a unit pixel mass from one position in $x$ to another in $\mathcal{I}_r$. In the discrete case, the classic optimal transport distance can be written as a linear optimization problem $\min_{\mathbf{P}\in\mathcal{U}} \langle\mathbf{D}, \mathbf{P}\rangle$, where $\mathcal{U} \coloneqq \{\mathbf{P} \in \mathbb{R}_+^{n \times n} | \mathbf{P1}_n = p_{x'}, \mathbf{P^\top1}_n = p_r \}$.

Further using the Lagrange multiplier, one can transform the problem into a regularized form as eq.~\ref{eq:lsp}. We name the optimal transportation loss, denoted as $\mathcal{L}_\mathrm{ot}$, serving as the minimum transportation effort on moving strokes from one location to another, and define it as the Sinkhorn distance in the following way:
\begin{equation}\label{eq:lsp}
\begin{aligned}
& \mathcal{L}_\mathrm{ot}(x', \mathcal{I}_r) = \langle\mathbf{D}, \tilde{\mathbf{P}}_\lambda \rangle \\
& \text{where } \tilde{\mathbf{P}}_\lambda = {\arg\min}_{\mathbf{P}\in\mathcal{U}} \langle\mathbf{D}, \mathbf{P} \rangle - \dfrac{1}{\lambda} E(\mathbf{P}) \\
& \text{and } E(\mathbf{P}) \coloneqq -\sum_{i,j=1}^n \mathbf{P}_{i,j} \log \mathbf{P}_{i,j}
\end{aligned}
\end{equation}

\noindent\textbf{Style-Preserving Loss $\mathcal{L}_\mathrm{sp}$.}
The optimized transport loss can be seamlessly integrated into the parameter search pipeline and optimized alongside other losses. Finally, we define the total loss objection as a combination of the optimal transport loss and the pixel $\ell_2$ loss with the DDIM inversion~\cite{song2021ddim} sample:
\begin{equation}
\mathcal{L}_\mathrm{sp} = \lambda_\mathrm{ot}\mathcal{L}_\mathrm{ot} + \lambda_{\ell_2}\mathcal{L}_{\ell_2}
\end{equation}
where $\lambda_\mathrm{ot}$ and $\lambda_{\ell_2}$ are weighting factors that balance the two loss terms.

\section{Implementation Details}
\label{supp:implementation}
\noindent Our method accepts a textual prompt to express semantics and a reference image to control the style. It is based on an optimization-based vector graphics synthesis pipeline~\cite{xing2024diffsketcher} with a differentiable rasterizer $\mathcal{R}$~\cite{li2020differentiable}, and style transfer methods in pixel space, InstantStyle~\cite{wang2024instantstyle} and StyleAligned~\cite{hertz2024style}.
As aforementioned in our manuscript in Sec.~\ref{subsec:stroke_extract}, We define the desired vector graphic $\mathcal{S}$ as a collection of $n$ vector strokes $\mathcal{S} = \{E_i\}_{i=1}^n$, where number of strokes $n$ is determined by the user.
These strokes serve as the set of parameters for the optimization process. If the strokes in the style image are short, users may choose to define a greater number of strokes; conversely, fewer strokes may be defined if the strokes are longer. For paintings by Vincent van Gogh, we typically set this number to approximately 3,000. For imitation learning, as mentioned in Sec.~\ref{subsec:svg_rendering}, we produce a vectorized version of the reference image $\mathcal{I}_s'$ with the optimization step $N_i=250$.

Furthermore, given a text prompt, we employ the Stable Diffusion XL~\cite{podell2023sdxl} as a pre-trained diffusion model, along with a DDIM inversion process~\cite{song2021ddim} with 50 steps and the CFG weight $\omega=7.5$, to sample a raster image. The parameters of SDXL are kept frozen in the optimization process.
During the optimization process, we optimize the stylized SVG for 2,000 steps, with the learning rate for the control point at 1.0, for width at 0.1, and for color at 0.05. We use the style-preserving loss, and both the $\lambda_\mathrm{ot}$ and $\lambda_{\ell_2}$ coefficients are set to 1.0.

Throughout the optimization process, VectorPainter requires 16GB memory for SDXL~\cite{podell2023sdxl} to sample, and only 3GB for the rest of the process. Our experiments were conducted on a single Tesla V100 GPU. The whole generation process, including DDIM inversion and optimization, takes less than 1 hour to complete to generate a stylized vector graphic consisting of approximately 3,000 strokes, about 30 minutes for 800 strokes, and approximately 10 minutes for 100 strokes.
Overall, the time consumption is comparable to other optimization-based methods.

\section{Stroke Style Extraction Algorithm}
\label{supp:stroke_extraction}
We summarize the Stroke Style Extraction Algorithm in Alg.~\ref{algo:supp_stroke_extraction}. The algorithm consists of two main steps: \textit{stroke extraction} (lines 1–11) and \textit{stroke vectorization} (lines 12–16). During the stroke extraction step, strokes are extracted from each segmented region after super-pixel segmentation, and the control points, color, and width attributes of each stroke are obtained. In the stroke vectorization step, an imitation strategy is applied to produce a vectorized version of the reference image, enhancing the style of the extracted strokes.

\section{Additional Qualitative Results}
\label{supp:more}
\begin{figure*}
    \centering
    \includegraphics[width=\textwidth]{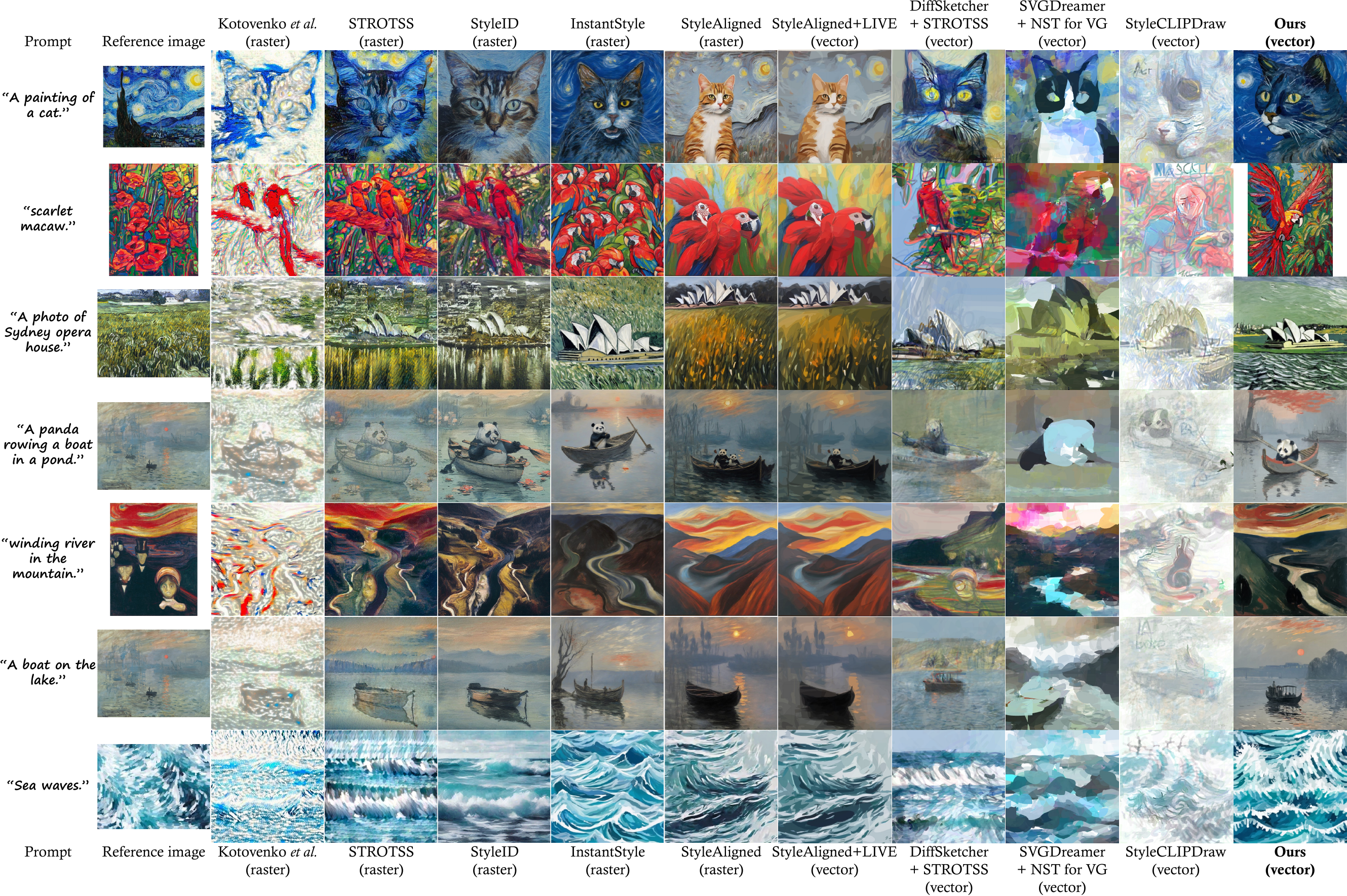}
    \caption{More Qualitative Results of our VectorPainter.}
    \label{fig:supp_compare}
\end{figure*}
In Fig.~\ref{fig:supp_compare}, we present additional results generated by VectorPainter. These examples demonstrate the model’s capability to produce stylized SVGs that not only maintain the semantic integrity of the textual prompt but also maintain the style of reference image with high visual quality.

\section{User Study}
\label{supp:user_study}
\begin{table}
\centering
\caption{Human Evaluation.}
\label{tab:user_study}
\resizebox{\linewidth}{!}{
\begin{tabular}{c|cc}
    \toprule
    Method & Text-Alignment$\downarrow$ & Style-Preservation$\downarrow$ \\
    \midrule
    Kotovenko \textit{et al.} (raster) & 6.10 & 5.56 \\
    STROTSS (raster) & 2.84 & 3.60 \\
    StyleID (raster) & 3.48 & 3.20 \\
    InstantStyle (raster) & 4.14 & 4.10 \\
    StyleAligned (raster) & 5.52 & 5.02 \\
    \midrule
    StyleAligned + LIVE (vector) & 5.76 & 5.98 \\
    DiffSketcher + STROTSS (vector) & 7.40 & 7.92 \\
    SVGDreamer + VectorNST (vector) & 8.06 & 8.38 \\
    StyleCLIPDraw (vector) & 9.60 & 8.98 \\
    \textbf{Ours (vector)} & \textbf{2.10} & \textbf{2.26} \\
    \bottomrule
\end{tabular}
\vspace{-2em}
}
\end{table}
We conducted a user study to compare our method with nine baseline methods in terms of text alignment and style preservation.
Given the limited time available, we used 10 text prompts and reference image combinations for evaluation, which are shown in Fig.~\ref{img:compare} and Fig.~\ref{fig:supp_compare}. 
12 participants were recruited to participate in this user study. 
Each participant was shown 5 combinations and the corresponding results from our method alongside the baseline methods.
Note that all results were displayed in random order, and we did not disclose which method was used for each result, nor did we specify whether the images were vector or raster graphics. Participants were asked to rank each set of images based on text alignment and style preservation, assigning a score from 1 (best) to 10 (worst).
In total, we collected 240 rankings from 12 users and then calculated the average ranking across all participants to determine each method’s overall score. The results are presented in Table~\ref{tab:user_study}. 
It is clear that our method significantly outperformed the other approaches.

\section{Limitations}
\label{supp:limitations}
Our method has a few limitations that need to be considered. Firstly, since our model is based on differentiable rendering, it can take considerable time to synthesize a vector graphic, which also exists in other optimization-based methods. Secondly, our method is based on the premise that the style of the reference image can be captured through the stroke style. However, when the image lacks a distinct stroke style (\textit{e.g.} non-painting works) or the stroke style is subtle (\textit{e.g.} watercolor), the performance of our method may diminish.

\end{document}